\documentclass[a4paper]{scrartcl}
\usepackage[latin1]{inputenc}
\usepackage[english]{babel}
\usepackage{mathptmx}
\usepackage{complexity}
\usepackage{amsmath}
\usepackage{amssymb}
\usepackage{amsfonts}
\usepackage{graphicx}
\usepackage[ruled,noline,linesnumbered,noend]{algorithm2e}
\usepackage[round]{natbib}
\usepackage{tikz}
\usepackage[all]{xy}
\usepackage{multirow}
\newcommand{\minimize}{minimize}
\newcommand{\subjectto}{subject to}
\newcommand{\minproblem}{\minproblemplain}
\newcommand{\minproblemplain}[3][]{
  \begin{align}
    \text{#1}\textbf{\minimize}\qquad & #2\\
    \textbf{\subjectto}\qquad & #3
  \end{align}
}
\newcommand{\maxproblem}{\@ifstar\maxproblemstar\maxproblemplain}
\newcommand{\maxproblemplain}[3][]{
  \begin{align}
    \text{#1}\textbf{\maximize}\qquad & #2\\
    \textbf{\subjectto}\qquad & #3
  \end{align}
}
\providecommand{\floor}[1]{\ensuremath{\left\lfloor #1\right\rfloor}}

\providecommand{\R}{\ensuremath{\mathbb R}}

\def\argmin{\mathop{argmin}}

\begin{document}

  \vspace*{4cm}
  
  \begin{center}
  \LARGE
  \textbf{Exact and Heuristic Methods for the Assembly Line Worker
    Assignment and Balancing Problem}
  \end{center}

  \begin{center}
    \bigskip
    \small
    Leonardo Borba, Marcus Ritt \\
    \{lmborba,mrpritt\}@inf.ufrgs.br \\
    Programa de Pós Graduação em Computação -- PPGC \\
    Instituto de Informática \\
    Universidade Federal do Rio Grande do Sul \\

    \bigskip
    Technical Report Number: 368 \\
    Universidade Federal do Rio Grande do Sul

  \end{center}

\vspace*{1cm}
\begin{abstract}
  \centerline{\textbf{Abstract}}
  In traditional assembly lines, it is reasonable to assume that task execution times are the same for each
  worker. However, in sheltered work centres for disabled this assumption is not valid: some workers may
  execute some tasks considerably slower or even be incapable of executing them. Worker heterogeneity leads to
  a problem called the assembly line worker assignment and balancing problem (ALWABP). For a fixed number of
  workers the problem is to maximize the production rate of an assembly line by assigning workers to stations
  and tasks to workers, while satisfying precedence constraints between the tasks.

  This paper introduces new heuristic and exact methods to solve this problem. We present a new MIP model,
  propose a novel heuristic algorithm based on beam search, as well as a task-oriented branch-and-bound
  procedure which uses new reduction rules and lower bounds for solving the problem. Extensive computational
  tests on a large set of instances show that these methods are effective and improve over existing ones.
\end{abstract}

\section{Introduction}
\label{sec:intro}

\par
The Universal Declaration of Human Rights states that ``everyone has the right to work, to free choice of
employment, to just and favourable conditions of work and to protection against
unemployment''~\citep{UN1948}. Despite this, low employment rates still demonstrate the lack of job
opportunities for more than 785 million persons with disabilities, including 110 million with a severe
deficiency degree, due to factors like prejudices and absence of appropriate technical
preparation~\citep{2003OECD}. This deficit lead to the creation of programs for the social inclusion of
persons with disabilities. Some of them concern their qualification~\citep{Organization2011}, while others
ensure opportunities by quota laws~\citep{Lobato2009}. Countries like Spain, Japan and Switzerland merged
these two forms by creating Sheltered Work Centres for Disabled (SWDs)~\citep{Chaves:2009:MeHiBu}, which
employ mainly persons with disabilities and provide training and a first job opportunity for
them~\citep{Miralles2007}. SWDs are not-for-profit industries applying all revenues in improvements for the
company and the creation of new jobs.

\citet{Miralles2007} have shown that using assembly lines in SWDs has advantages, because the division of work
into smaller tasks can effectively hide the differences among the workers. Furthermore, the execution of
repetitive tasks, when properly assigned, can be an excellent therapeutic treatment for workers with
disabilities. Traditional approaches to the optimization of assembly lines assume that the workers have
similar abilities and are capable of executing the tasks in the same time. The most basic model of this kind
is the Simple Assembly Line Balancing Problem (SALBP), which has been extensively studied in the
literature~\citep{Scholl2006}. Several authors have considered stochastic models of assembly lines, where task
times may vary, and remedial actions are taken if the cycle time is exceeded at some
station~\citep{Silverman.Carter/1986,Kottas.Lau/1976,Lyu/1997}. In this paper we are not directly concerned
with varying task times of a single worker, but with the case of SWDs, where the workers need different times
to execute the tasks, or may even be incapable of executing some of them. To model such a situation,
\citet{miralles2008branch} proposed the Assembly Line Worker Assignment and Balancing Problem (ALWABP), which
assigns tasks to different workers and these workers to the workstations.

\subsection{Problem Definition}

\par
Let $S$ be a set of stations, $W$ be a set of workers, $|W|=|S|$, and $T$ be a set of tasks. Each workstation
$s \in S$ is placed along a conveyor belt and is assigned to exactly one worker $w \in W$, which is
responsible for executing a subset of tasks $x_{w} \subseteq T$. The tasks are partially ordered, and we
assume that the partial order is given by a transitively reduced directed acyclic graph $G(T,E)$ on the tasks,
such that for an arc $(t,t')\in E$ task $t$ precedes task $t'$. Therefore, the station that executes task $t$
cannot be placed later than that of task $t'$ on the conveyor belt. The execution time of task $t$ for worker
$w$ is $p_{tw}$. If a worker $w$ cannot execute a task $t$, $p_{tw}$ is set to $\infty$.

\par
The total execution time of worker $w$ is $D_w=\sum_{t\in x_w} p_{wt}$. The cycle time $C$ of the line is
defined by the maximum total execution time $\max_{w\in W} D_w$. In assembly line balancing, a problem of
\emph{type 1} aims to reduce the number of stations for a given cycle time. Since in SWDs the goal is to
include all workers, our problem is of \emph{type 2}, and aims to minimize the cycle time for a given number
of stations and the same number of workers. A valid solution is an assignment of workers to stations together
with an assignment of tasks to workers that satisfies the precedence constraints.

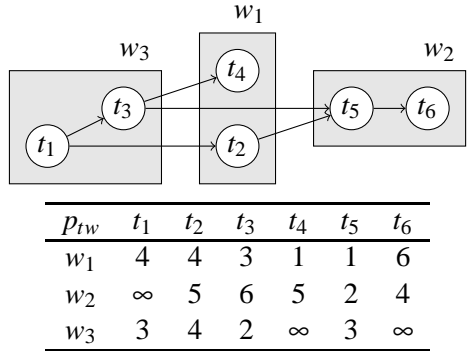
\begin{figure}[t]
  \centering
  \begin{tikzpicture}[vertex/.style={draw,circle,inner sep=0.4ex,fill=white}]
    \draw[fill=black!10] (0.5,-0.5) rectangle (2.5,1) node[above left] { $w_3$ };
    \draw[fill=black!10] (3,-0.5) rectangle (4,1.5) node[above left] { $w_1$ };
    \draw[fill=black!10] (4.5,0) rectangle (6.5,1) node[above left] { $w_2$ };
    \node[vertex] (n1) at (1,0) { $t_1$ };
    \node[vertex] (n2) at (3.5,0) { $t_2$ };
    \node[vertex] (n3) at (2,0.5) { $t_3$ };
    \node[vertex] (n4) at (3.5,1) { $t_4$ };
    \node[vertex] (n5) at (5,0.5) { $t_5$ };
    \node[vertex] (n6) at (6,0.5) { $t_6$ };
    \draw[->] (n1) -- (n2);
    \draw[->] (n1) -- (n3);
    \draw[->] (n3) -- (n4);
    \draw[->] (n3) -- (n5);
    \draw[->] (n2) -- (n5);
    \draw[->] (n5) -- (n6);
  \end{tikzpicture}

  \medskip
  \begin{tabular}{ccccccc}
    \hline
    $p_{tw}$ & $t_1$ & $t_2$ & $t_3$ & $t_4$ & $t_5$ & $t_6$\\
    \hline
    $w_1$ & 4 & 4 & 3 & 1 & 1 & 6\\
    $w_2$ & $\infty$ & 5 & 6 & 5 & 2 & 4\\
    $w_3$ & 3 & 4 & 2 & $\infty$ & 3 & $\infty$\\
    \hline
  \end{tabular}

  \caption{Example of an ALWABP instance and an assignment of tasks to workers (in grey). Upper part:
    precedence constraints among the tasks. Lower part: task execution times.}
  \label{fig:example}
\end{figure}

\par
Figure \ref{fig:example} shows an example of an ALWABP-2 instance. For the assignment given in the figure, we
have $D_{w_1}=5$, $D_{w_2}=6$, $D_{w_3}=5$, and a cycle time of $C = \max{\{D_{w_1},D_{w_2},D_{w_3}\}} = 6$.

\subsection{Related Work}

\par
The majority of the publications on the ALWABP-2 is dedicated to the application of meta-heuristics to find
approximate solutions to the problem. Two clustering search methods were proposed by
\citet{chaves2007clustering,Chaves2009c}, which were outperformed on large instances by a tabu search of
\citet{moreira2009minimalist}. \citet{Blum.Miralles/2011} proposed an iterated beam search based on the
station-oriented branch-and-bound procedure of~\citet{miralles2008branch}. Later, \citet{Moreira2012Simple}
used a constructive heuristic with various combinations of priority rules to produce initial solutions for a
genetic algorithm (GA). \citet{Mutlu2013} developed an iterated GA that produces valid orders of tasks and
applies iterated local search to attribute the tasks in the selected order to the workers.

\par
To the best of our knowledge, the only exact method for the ALWABP-2 is the branch-and-bound procedure
of~\citet{miralles2008branch}. It embeds a station-oriented, depth-first branch-and-bound search in a linear
lower bound search for the optimal cycle time, and is limited to very small instances.

\subsection{Structure of the paper}

\par
In Section~\ref{sec:definition} we introduce a new MIP model for the ALWABP-2. In
Section~\ref{sec:lowerbounds} we present several lower bounds for the problem. A new heuristic for ALWABP-2 is
proposed in Section~\ref{sec:upperbounds}. In Section~\ref{sec:branchandbound} we present a task-oriented
branch-and-bound method for solving the problem exactly. Computational results are presented and analyzed in
Section \ref{sec:results}. We conclude in Section~\ref{sec:conclusion}.

\section{Mathematical formulation}
\label{sec:definition}

\par
In this section we will present a new mixed-integer model for the ALWABP-2. Currently, the only model used in
the literature, called $M_1$ here, is the one proposed by \citet{miralles2008branch}. It has
$O(|T|\,|W|\,|S|)$ variables, and $O(|T|+|E|+|W|\,|S|)$ constraints. In the following we will use the notation
defined in Table~\ref{table:notation}.
\begin{table}
  \center
  \caption{Notation for ALWABP-2.}\medskip
  \label{table:notation}
  \begin{tabular}{p{95pt} p{270pt}}\hline
    $S$                 & set of stations;                                                   \\
    $W$                 & set of workers;                                                    \\
    $T$                 & set of tasks;                                                      \\
    $G(T,E)$            & transitively reduced precedence graph of tasks;                    \\
    $G^*(T,E^*)$        & transitive closure of graph $G(T,E)$;                              \\
    $p_{tw} $           & execution time of task $t$ by worker $w$;                          \\
    $A_w \subseteq T$   & set of tasks feasible for worker $w$;                              \\
    $A_t \subseteq W$   & set of workers able to execute task $t$;                           \\
    $P_t$ and $F_t$     & set of direct predecessors and successors of task $t$ in $G$; \\
    $P^*_t$ and $F^*_t$ & set of all predecessors and successors of task $t$ in $G^*$;       \\
    $C \in \mathbb{R}$  & cycle time of a solution.                                      \\\hline
  \end{tabular}
\end{table}

\subsection{Formulation with two-index variables}

Our formulation is based on the observation that it is sufficient to assign tasks to workers and to guarantee
that the directed graph over the workers, induced by the precedences between the tasks, is acyclic. Therefore
our model uses variables $x_{wt}$ such that $x_{wt}=1$ if task $t\in T$ has been assigned to worker $w\in W$,
and $d_{vw}$ such that $d_{vw}=1$ if worker $v\in W$ must precede worker $w\in W$. In this way, we obtain a
model $M_2$ as follows:
\minproblem{C,\label{i:1}}{
  \sum\limits_{t \in A_w} p_{tw}\, x_{wt} \leq C, & \forall w\in W,\label{i:3}\allowdisplaybreaks\\
  & \sum\limits_{w\in A_t} x_{wt} = 1, & \forall t\in T,\label{i:2}\allowdisplaybreaks\\
  & d_{vw} \geq x_{vt} + x_{wt'} - 1, & \forall (t,t') \in E, v \in A_t, w \in A_{t'} \setminus \{v\},\label{i:6}\allowdisplaybreaks\\
  & d_{uw} \geq d_{uv} + d_{vw} - 1, &  \forall \{u,v,w\}\subseteq W,|\{u,v,w\}|=3,\label{i:trans}\allowdisplaybreaks\\
  & d_{vw} + d_{wv} \leq 1, & \forall v \in W, w \in W \setminus\{v\},\label{i:6.2}\allowdisplaybreaks\\
  & x_{wt}\in\{0,1\}, & \forall w\in W, t\in A_w,\label{i:7}\allowdisplaybreaks\\
  & d_{vw}\in\{0,1\}, & \forall v \in W, w\in W\setminus\{v\},\label{i:8}\allowdisplaybreaks\\
  & C\in\R\label{i:9}\textrm{.}}

\par
Constraint \eqref{i:3} defines the cycle time $C$ of the problem. Constraint \eqref{i:2} ensures that every
task is executed by exactly one worker. The dependencies between workers are defined by
constraint~\eqref{i:6}: when a task $t$ is assigned to worker $v$ and precedes another task $t'$ assigned to a
different worker $w$, worker $v$ must precede worker $w$. Constraints~\eqref{i:trans} and \eqref{i:6.2}
enforce transitivity and anti-symmetry of the worker dependencies. As a consequence of these constraints, the
workers of a valid solution can always be ordered linearly.

\subsection{Continuity constraints}

We can strengthen the above model by the following observation: if two tasks $i$ and $k$ are assigned to the
same worker $w$, then all tasks $j$ that are simultaneously successors of $i$ and predecessors of $k$ should
also be assigned to $w$. These \emph{continuity constraints} generalize constraints proposed by
\citet{Peeters.Degraeve/2006} for single station loads in the SALBP to several stations:
\begin{align}
  & x_{wj} \geq x_{wt} + x_{wk} - 1, & \forall i\in T,j\in F_i^*,k\in F_j^*,w \in A_i \cap A_j \cap A_k\mathrm{.} \label{nc:1}
\end{align}
Similarly, if task $i$ is assigned to worker $w$, but some successor (predecessor) $j$ of $i$ is unfeasible
for $w$, then no successor (predecessor) of $j$ can be assigned to $w$. This justifies the constraints
\begin{align}
  & x_{wk} + x_{wi} \leq 1, & \forall i\in T,j\in F_i^*,k\in F_j^*,w \in A_i \cap (T\setminus A_j) \cap A_k\mathrm{.}\label{nc:2}
\end{align}
Let model $M_3$ be model $M_2$ with additional constraints \eqref{nc:1} and \eqref{nc:2}. Model $M_3$ has
$O(|W|(|T|+|W|))$ variables, and $O(|E^*||T||W|+|W|^3+|E||W|^2)$ constraints, i.e.~it has less variables but
more constraints than $M_1$. As will be seen in Section~\ref{sec:results} model $M_3$ gives significantly
better bounds than $M_1$.

\section{Lower bounds}
\label{sec:lowerbounds}

Lower bounds for ALWABP-2 can be obtained by different relaxations of the problem. In this section we discuss
relaxations of the mixed-integer model presented above, as well as relaxations to SALBP-2 and $R\mid\mid
C_\mathrm{max}$.

\subsection{Relaxation to SALBP-2}
\label{sec:salbp2}

If we relax the task processing times to their minimum $p^-_t = \min{\{p_{tw} \mid w \in W\}}$, ALWABP-2
reduces to SALBP-2. Therefore, all valid lower bounds for SALBP-2 apply to this relaxation. In particular, we
use the lower bounds
\begin{align*}
  LC_{1} & = \max\left\{\max\{p^-_t \mid t \in T\},\biggl\lceil\sum_{t \in T}{(p^-_t)} / |S|\biggr\rceil\right\}\quad\text{and}\\
  LC_{2} & = \max{\left\{ \sum\limits_{0\leq i\leq k}{p^-_{k|S| + 1 - i}}~ \Big{|} ~1 \leq k \leq \left\lfloor{\frac{|T| - 1}{|W|}}\right\rfloor\right\}}\mathrm{.}
\end{align*}
(The bound $LC_2$ supposes that the tasks are ordered such that $p^{-}_1\geq\cdots\geq p^{-}_{|T|}$.)
We further use the SALBP-2 bounds on the earliest and latest possible station of task $t$ for a given cycle
time $C$
\begin{eqnarray}
  \label{eq:earliest}
  E_t(C) = \left\lceil{\biggl({\sum_{j \in P^*_i}{p^-_j + p^-_t}}\biggr) / C }\right\rceil \quad\text{and}\\
  \label{eq:latest}
  L_t(C) = |S| + 1 - \left\lceil{\biggl({\sum_{j \in F^*_i}{p^-_j + p^-_t}}\biggr) / C }\right\rceil
\end{eqnarray}
to obtain the lower bound
$LC_3$, defined as the smallest cycle time $C$ such that $E_t(C) \leq L_t(C)$ for all $t \in T$. For more
details on these bounds we refer the reader to the survey of \citet{Scholl.Becker/2006}.

\subsection{Relaxation to $R\mid\mid C_\mathrm{max}$}

By removing the precedence constraints the ALWABP-2 reduces to the problem of minimizing the makespan of the
tasks on unrelated parallel machines ($R\mid\mid C_\mathrm{max}$), which itself is an NP-hard problem. Several
effective lower bounds for $R\mid\mid C_\mathrm{max}$ have been proposed by \citet{Martello1997}. Their lower
bounds $L_1$ and $L_2$ are obtained by Lagrangian relaxation of the cycle time constraints~\eqref{i:3} and the
assignment constraints~\eqref{i:2}, respectively. \citet{Martello1997} further propose an additive improvement
that can be applied to $L_1$ to obtain a bound $L_1^a\geq L_1$, as well as an improvement by cuts on
disjunctions, that may be applied to $L_1^a$ and $L_2$ to obtain lower bounds $\overline{L}_1^a\geq L_1^a$ and
$\overline{L}_2\geq L_2$.

\subsection{Linear relaxation of ALWABP-2 models}

Bounds obtained from linear relaxations of integer models for the SALBP-2 are usually weaker than the SALBP-2
bounds of Section~\ref{sec:salbp2}. However, the relaxation to minimum task execution times weakens the
SALBP-2 bounds considerably. Therefore, the linear relaxations of model $M_3$ provides a useful lower bound
for the ALWABP-2~\citep{Moreira2012Simple}.

\section{Heuristic search procedure}
\label{sec:upperbounds}

In this section, we describe a heuristic algorithm IPBS for the ALWABP-2. It systematically searches for a
small cycle time by trying to solve the feasibility problem ALWABP-F for different candidate cycle times from
an interval ending at the current best upper bound. For each candidate cycle time $C$, a probabilistic beam
search tries to find a feasible allocation.

\subsection{Probabilistic beam search for the ALWABP-F}

The basis for the probabilistic beam search is a station-based assignment procedure, which assigns tasks in a
\emph{forward} manner station by station.  For each station it repeatedly selects an available task, until no
such task has an execution time less than the idle time of the current station. A task is \emph{available} if
all its predecessors have been assigned already. If there are several available tasks the highest priority
task as defined by a prioritization rule is assigned next. The procedure succeeds if an assignment using at
most the available number of stations is found. Station-based procedures can be also applied in a
\emph{backward} manner, assigning tasks whose successors have been assigned already. For this it is sufficient
to apply a forward procedure to an instance with reversed dependencies. For the ALWABP we additionally have to
decide which worker to assign to the current station. This is accomplished by applying the task assignment
procedure to all workers which are not yet assigned, and then choosing the best worker for the current station
by a worker prioritization rule.

The probabilistic beam search extends the station-oriented assignment procedure in two aspects. First, when
assigning tasks to the current station, it randomly chooses one of the available tasks with a probability
proportional to its priority. Second, it applies beam search to find the best assignment of workers and their
corresponding tasks.

Beam search is a truncated breadth-first tree search procedure~\citep{beamharpy,PENG1988}. When applied to the
ALWABP-F, it maintains a set of partial solutions called the \emph{beam} during the station-based
assignment. The number of solutions in the beam is called its \emph{width} $\gamma$. Beam search extends a
partial solution by assigning each available worker to the next station, and for each worker, chooses the
tasks to execute according to the above probabilistic rule. For each worker this is repeated several times, to
select different subsets of tasks. The number of repetitions is the beam's \emph{branching factor} $f$. Among
all new partial solutions the algorithm selects those of highest worker priority to form the new beam. The
number of solutions selected is at most the beam width.

Task and worker prioritization rules are important for the efficacy of station-oriented assignment
procedure. \citet{Moreira2012Simple} compared the performance of 16 task priority rules and three worker
prioritization rules for the \mbox{ALWABP-2}. We have chosen the task priority rule $MaxPW^-$ and the worker
priority rule $MinRLB$, which have been found to produce the best results in average for the problem. The task
prioritization rule $MaxPW^-$ gives preference to tasks with larger minimum positional weight $pw^-_t = p^-_t
+ \sum_ {t' \in F^*_t}{p^-_{t'}}$. The worker prioritization rule $MinRLB$ gives preference to workers with
smaller \textit{restricted lower bound} $\sum_{t \in T_u}{p^-_t(W_u)/|W_u|}$, where $p^-_t(W') = \min_{w \in
  W'}{p_{tw}}$ with the set $W_u \subseteq W$ corresponding to the unassigned workers and $T_u \subseteq T$ to
the set of unassigned tasks of a partial assignment. Before computing $MinRLB$ we apply to each partial
solution the logic of the continuity constraints \eqref{nc:1} and \eqref{nc:2} to strengthen the bound. If
tasks $i$ and $k$ have been assigned already to some worker $w$, we also assign all tasks succeeding $i$ and
preceding $k$ to $w$. Similarly, if $i$ has been assigned to $w$ and some successor (predecessor) $j$ of $i$
is infeasible for $w$ we set $p_{kw}=\infty$ for all successors (predecessors) $k$ of $j$. The probabilistic
beam search is shown in Algorithm~\ref{alg:beamsearch}.

\begin{algorithm}
  \footnotesize

  \SetInd{0.3em}{0.5em}

  \caption{Probabilistic beam search}
  \label{alg:beamsearch}
  \SetInd{0.3em}{0.5em}
  \SetKwData{Left}{left}\SetKwData{This}{this}\SetKwData{Up}{up}
  \SetKwFunction{Union}{Union}\SetKwFunction{FindCompress}{FindCompress}
  \SetKwInOut{Input}{input}\SetKwInOut{Output}{output}
  \Input{A set of stations $S$, a candidate cycle time $C$, a beam width $\gamma$ and a beam factor $f$.}
  \Output{A valid assignment or ``failed'' if no valid assignment could be found.}

  $B \gets \{\emptyset\}$\tcc*{current set of partial assignments}
  \For{$k \in S$} {
    $B' \gets \emptyset$\;
    \For{$s \in B$}{
      \For{$f$ times}{
        \For{all unassigned workers $w \in W$} {
          $s' \gets s $ concatenated with a new empty station $k$\;
          \While{there are available tasks $P$ that do not overload the current station}{
            select a task $t \in P$ with probability proportional to $MaxPW^-(t)$\;
            assign $t$ to station $k$ in $s'$\;
          }
          \lIf{all tasks in $T$ are assigned in $s'$}{
            \Return{Solution $s'$}\;
          } \lElseIf{$|B'| < \gamma$}{
            $B' \gets B' \cup \{s'\}$\;
          } \Else {
            $o \gets  \argmin{\{MinRLB(o')~|~o' \in B'\}}$\;
            \lIf{$MinRLB(s') > MinRLB(o)$}{
              $B' \gets B' \cup \{s'\} \setminus \{o\}$\;
            }
          }
        }
      }
    }
    $B \gets B'$\;
    \Return{``failed''}\;
  }
\end{algorithm}

\subsection{The interval search method IPBS}

An upper bound search starts from a known feasible cycle time and tries to reduce it iteratively. A common
strategy is to reduce it successively by one and to try to find a better feasible solution by some heuristic
algorithm. However, it is well known that heuristic assignment procedures are not monotone, i.e., they may
find a feasible solution for some cycle time but not for larger cycle times. To overcome this, we propose to
modify the upper bound search to examine an interval of cycle times ending at the current best upper bound. If
the current lower and upper bounds on the cycle time are $\underline C$ and $\overline C$, the upper bound
search will try to find a feasible solution for all cycle times between $\max\{\underline C,\floor{p\overline
  C}\}$ and $\overline C-1$ for a given factor $p\in(0,1)$ and update $\overline C$ to the best cycle time
found, if any. Otherwise, the upper bound search continues with the same interval. Since the beam search is
probabilistic this may produce a feasible solution in a later trial. The interval search depends on three
parameters: the minimum search time $t_\mathrm{min}$, the maximum search time $t_\mathrm{max}$ and the maximum
number of repetitions $r$. The search terminates if the cycle time found equals the lower bound, or if the
maximum time or the maximum number of repetitions are exceeded, but not before the minimum search time has
passed.

Initially, the value of $\underline{C}$ is set to the best of all lower bounds presented in
Section~\ref{sec:lowerbounds}. The initial upper bound $\overline{C}$ is determined by an single run of the
beam search with a beam factor of one.

\subsection{Improvement by local search}

A local search is applied to the results found by interval search method. It focuses on \textit{critical
  stations} whose load equals the cycle time of the current assignment. It tries to remove tasks from a
critical station in order to reduce the cycle time. Since there can be multiple critical stations, a move is
considered successful if it reduces the number of critical stations. The local search applies the following
four types of moves, until the assignment cannot be improved any more.

\begin{enumerate}
\item A \textit{shift} of a task from a critical station to another station.
\item A \textit{swap} of two tasks. At least one of the tasks must be on a critical station.
\item A sequence of two shift moves. Here the first shift move is allowed to produce a worse result than the
  initial assignment.
\item A swap of workers between two stations without reassigning the tasks.
\end{enumerate}

\section{Task-oriented branch-and-bound algorithm}
\label{sec:branchandbound}

In this section we propose a branch-and-bound algorithm for ALWABP-2 using the bounds and the heuristic
presented in the previous sections.

The algorithm first computes a heuristic solution by running the probabilistic beam search. It also applies
the lower bounds $LC_1,LC_2,LC_3,M_3,\overline{L}_1^a,\overline{L}_2$ at the root node to obtain an initial
lower bound.

If the solution cannot be proven optimal at the root node, the algorithm proceeds with a depth-first
search. In branch-and-bound algorithms for assembly line balancing two branching strategies are common.  The
\textit{station-oriented} method proceeds by stations and branches on all feasible maximal loads for the
current station, while the \emph{task-oriented} method, proceeds by tasks and branches on all possible
stations for the current task. The most effective methods for SALBP use station-oriented branching. However,
for the ALWABP the additional worker selection substantially increases the branching factor of the
station-oriented approach. A \emph{worker-oriented} strategy, on the other hand, has to consider much more
station loads, since all subsets of unassigned tasks which satisfy the continuity constraints~\eqref{nc:1} are
candidates to be assigned to a worker. Therefore, we use a task-oriented branching strategy.

The proposed task-oriented method executes the recursive procedure shown in Algorithm
\ref{alg:branchandbound}. At each new node it applies the lower bounds $LC_1,LC_2,LC_3,\overline{L}_1^a$
(line~\ref{line:newlb}), since the lower bounds $M_3$ and $\overline{L}_2$ are too slow to be applied during
the search, although they obtain the best bounds. When a complete solution has been found, the algorithm
updates the incumbent (line~\ref{line:initsolv}). Otherwise, it selects an unassigned task $t$
(line~\ref{line:selecttask}) and assigns it to all feasible workers (loop in
lines~\ref{line:sortworkers}--\ref{line:unsetassignment}).

\begin{algorithm}
  \small
  \caption{$\mathbf{branchTasks}(llb,A)$}
  \label{alg:branchandbound}
  \SetInd{0.3em}{0.5em}
  \SetKwData{Left}{left}\SetKwData{This}{this}\SetKwData{Up}{up}
  \SetKwFunction{Union}{Union}\SetKwFunction{FindCompress}{FindCompress}
  \SetKwInOut{Input}{Input}\SetKwInOut{Output}{output}
  \Input{An upper bound $gub$, a set $A\subseteq T$ of assigned tasks, and a local lower bound $llb$.}
  \BlankLine
  \nllabel{line:initsolv}\If{$A = T$}{
    \lIf{$llb < gub$}{$gub \leftarrow llb$}\;
    \Return{}
  }\nllabel{line:endsolv}
  \textbf{select a task } $t \in T \setminus A$ \nllabel{line:selecttask}\;
  \ForEach{$w \in W\mid \mathrm{assignmentIsValid}(t,w)$ \nllabel{line:sortworkers}}{
    \textbf{apply reduction rules}\nllabel{line:cuts}\;
    $newllb \gets \mbox{\textbf{lower bound with new assignment} } (t,w)$\nllabel{line:newlb}\;
    \If{$newllb < gub$}{
      \textbf{setAssignment}$(t,w)$\nllabel{line:setassignment}\;
      $\mathrm{branchTasks}(newllb,A\cup\{t\})$\;
      \textbf{unsetAssignment}$(t,w)$\nllabel{line:unsetassignment}\;
    }
  }
\end{algorithm}

For branching, the task with the largest number of infeasible workers is chosen in
line~\ref{line:selecttask}. A worker is considered infeasible, if the allocation of the task to the worker
creates an immediate cyclic worker dependency or the lower bound $LC_1$ after the assignment is at least the
value of the incumbent. In case of ties, the task with the largest lower bound is chosen, where the lower
bound of a task is the smallest lower bound $LC_1$ over its feasible workers. This rule gives preference to
tasks that tighten the lower bound early. Any remaining tie is broken by the task index. After the task has
been chosen, a branch is created for each valid worker. The branches are visited in order of non-decreasing
lower bounds. Again, ties are broken by the worker index.

\subsection{Valid assignments}

The algorithm maintains a directed graph $H$ on the set of workers to verify efficiently if the precedence
constraints are satisfied. It contains an edge $(w,w')$ if there is some task $t$ assigned to $w$ and another
task $t'$ assigned to $w'$, such that $(t,t')\in E^*$. The graph $H$ also contains all resulting transitive
edges. For a valid assignment of tasks, $H$ must be acylic. If this is the case, any topological sorting
defines a valid assignment of workers to stations. The procedure $\mathrm{assignmentIsValid}(t,w)$ verifies in
time $O(|T||W|)$ if the assignment of task $t$ to worker $w$ would insert an arc into $H$ whose inverse arc
exists already. Before branching to a new node, the procedure $\mathrm{setAssignment}(t,w)$ inserts such arcs
into $H$ and computes the new transitive closure in time $O(|T||W|)$. This is undone by
$\mathrm{unsetAssignment}(t,w)$ when backtracking. To speed up the selection of a task for branching, we do
not consider the violation of transitive dependencies in $H$ in line~\ref{line:selecttask}, but only the
creation of an immediate cyclic worker dependency, which results from inserting an edge $(w,w')$ for which
$(w',w)$ is already present. This can be tested in time $O(|P_t|+|F_t|)$.

\subsection{Reduction rules}
\label{subsection:reductionrules}

After a task $t$ has been assigned to a worker $w$, and before branching, we apply several more costly
reduction rules to strengthen the lower bounds (line~\ref{line:cuts}). First, we can set $p_{tw'}=\infty$ for
any $w'\neq w$. Second, we can enforce the continuity constraints \eqref{nc:1} and \eqref{nc:2}. An
application of \eqref{nc:1} may assign further tasks to $w$, and the application of \eqref{nc:2} may exclude
some tasks from being assigned to $w$ (whose execution time is set to $p_{t'w}=\infty$). Finally, we can
exclude a task-worker assignment $(t',w)$ if the total execution time $p_{tw} + p_{t'w} + \sum_{u \in
  i(t,t')}{p_{uw}}$ of the tasks $i(t,t') = (P^*_t \cap F^*_{t'}) \cup (F^*_t \cap P^*_{t'})$ between $t$ and
$t'$ is more than or equal to the current upper bound. These rules are repeatedly applied until no more tasks
can be assigned or excluded.

\section{Computational results}
\label{sec:results}

All algorithms were implemented in C++ and compiled with the GNU C compiler 4.6.3 with maximum
optimization. The MIP models and their linear relaxations were solved using the commercial solver CPLEX
12.3. The experiments were done on a PC with a 2.8 GHz Core i7 930 processor and 12 GB of main memory, running
a 64-bit Ubuntu Linux. All tests used only one core. Details of the results reported in this section are
available online \footnote{http://www.inf.ufrgs.br/algopt}.
\subsection{Test instances}

A set of $320$ test instances has been proposed by \citet{chaves2007clustering}. They are characterized by
five experimental factors: the number of tasks, the order strength (OS)\footnote{The order strength is number
  of precedence relations of the instance in percent of all possible relations $\binom{|T|}{2}$.}, the number
of workers ($|W|$), the task time variability (Var), and the percentage of infeasible task-worker pairs
(Inf). All factors take two levels, as shown in Table~\ref{tab:values}, defining $32$ groups of $10$
instances. The instances are based on the SALBP instances \emph{Heskia} (low $|T|$, low OS) , \emph{Roszieg}
(low $|T|$, high OS), \emph{Wee-mag} (high $|T|$, low OS), and \emph{Tonge} (high $|T|$, high OS). The first
worker of each instance executes task $t\in T$ in time $p_t$ of the corresponding SALBP instance, and the
remaining workers have an execution time randomly selected in the interval $[1,p_i]$ (low variability) or
$[1,2p_i]$ (high variability).

\begin{table}
  \caption{Instance characteristics. The $320$ instances are grouped by five two-level experimental factors
    into $32$ groups of $10$ instances.}
  \small
  \begin{center}
    \begin{tabular}{l c c} \noalign{\hrule height 1pt} \textbf{Factor} & \textbf{Low Level} & \textbf{High  Level} \\\noalign{\hrule height 1pt}
      Number of tasks $|T|$                                            & $25 - 28$          & $70 - 75$            \\
      Order strength (OS)                                              & $22\%$ - $23\%$    & $59\%$ - $72\%$      \\
      Number of workers $|W|$                                          & $|T| / 7$          & $|T| / 4$            \\
      Task time variability (Var)                                      & $[1,t_i]$          & $[1,2t_i]$           \\
      Number of infeasibilities (Inf)                                  & $10\%$             & $20\%$               \\\noalign{\hrule height 1pt}
    \end{tabular}
  \end{center}
  \label{tab:values}
\end{table}

\subsection{Comparison of lower bounds}

We first compare the strength of the lower bounds proposed in Section~\ref{sec:lowerbounds}. To compute the
lower bound $L_1$ we use the ascent direction method of \citet{velde1993duality}. This bound was improved to
$\overline{L}_1^a$, as proposed by \citet{Martello1997}. Their method applies a binary search for the best
improved bound, which is obtained by solving $|S|$ knapsack problems of capacity $C$ for each trial cycle time
$C$. Different from \citet{Martello1997} we solve the all-capacities knapsack problem by dynamic programming
only once and use the resulting table during the binary search. The knapsack problems that arise when
computing $L_2$ and $\overline{L}_2$ by subgradient optimization are solved similarly.

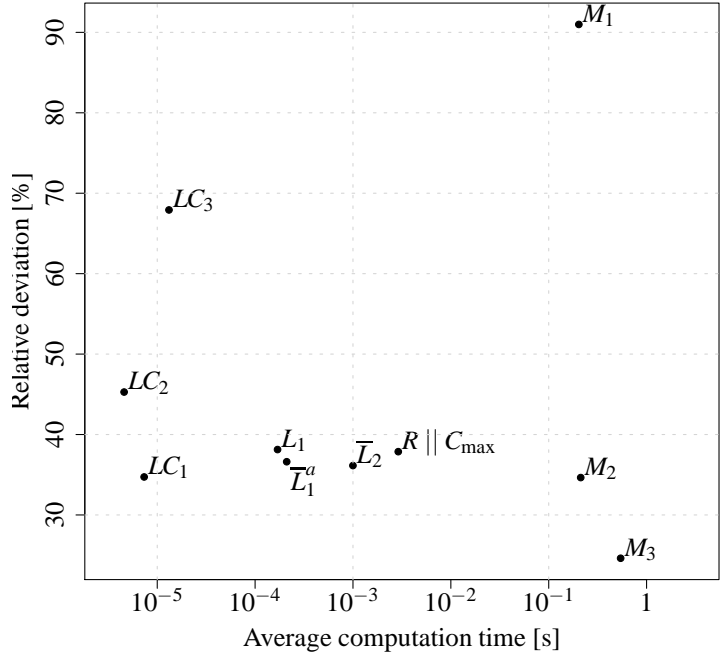
\begin{figure}[t!]
  \centering\small
  \begin{tikzpicture}[x=1pt,y=1pt,scale=0.55]
    \definecolor[named]{fillColor}{rgb}{1.00,1.00,1.00}
    \path[use as bounding box,fill=fillColor,fill opacity=0.00] (0,0) rectangle (505.89,505.89);
    \begin{scope}
      \path[clip] ( 49.20, 61.20) rectangle (480.69,456.69);
      \definecolor[named]{drawColor}{rgb}{0.00,0.00,0.00}
      \definecolor[named]{fillColor}{rgb}{0.00,0.00,0.00}

      \path[draw=drawColor,line width= 0.4pt,line join=round,line cap=round,fill=fillColor] (385.32,442.04) circle (  2.25);

      \path[draw=drawColor,line width= 0.4pt,line join=round,line cap=round,fill=fillColor] (186.57,142.05) circle (  2.25);

      \path[draw=drawColor,line width= 0.4pt,line join=round,line cap=round,fill=fillColor] (180.33,150.41) circle (  2.25);

      \path[draw=drawColor,line width= 0.4pt,line join=round,line cap=round,fill=fillColor] (231.62,139.42) circle (  2.25);

      \path[draw=drawColor,line width= 0.4pt,line join=round,line cap=round,fill=fillColor] ( 89.57,131.61) circle (  2.25);

      \path[draw=drawColor,line width= 0.4pt,line join=round,line cap=round,fill=fillColor] ( 75.94,189.86) circle (  2.25);

      \path[draw=drawColor,line width= 0.4pt,line join=round,line cap=round,fill=fillColor] (106.47,314.78) circle (  2.25);

      \path[draw=drawColor,line width= 0.4pt,line join=round,line cap=round,fill=fillColor] (386.60,131.19) circle (  2.25);

      \path[draw=drawColor,line width= 0.4pt,line join=round,line cap=round,fill=fillColor] (413.76, 75.85) circle (  2.25);

      \path[draw=drawColor,line width= 0.4pt,line join=round,line cap=round,fill=fillColor] (262.46,148.96) circle (  2.25);
    \end{scope}
    \begin{scope}
      \path[clip] (  0.00,  0.00) rectangle (505.89,505.89);
      \definecolor[named]{drawColor}{rgb}{0.00,0.00,0.00}

      \path[draw=drawColor,line width= 0.4pt,line join=round,line cap=round] ( 98.48, 61.20) -- (431.41, 61.20);

      \path[draw=drawColor,line width= 0.4pt,line join=round,line cap=round] ( 98.48, 61.20) -- ( 98.48, 55.20);

      \path[draw=drawColor,line width= 0.4pt,line join=round,line cap=round] (165.06, 61.20) -- (165.06, 55.20);

      \path[draw=drawColor,line width= 0.4pt,line join=round,line cap=round] (231.65, 61.20) -- (231.65, 55.20);

      \path[draw=drawColor,line width= 0.4pt,line join=round,line cap=round] (298.24, 61.20) -- (298.24, 55.20);

      \path[draw=drawColor,line width= 0.4pt,line join=round,line cap=round] (364.83, 61.20) -- (364.83, 55.20);

      \path[draw=drawColor,line width= 0.4pt,line join=round,line cap=round] (431.41, 61.20) -- (431.41, 55.20);

      \node[text=drawColor,anchor=base,inner sep=0pt, outer sep=0pt, scale=  1.00] at ( 98.48, 39.60) {${10}^{-5}$};

      \node[text=drawColor,anchor=base,inner sep=0pt, outer sep=0pt, scale=  1.00] at (165.06, 39.60) {${10}^{-4}$};

      \node[text=drawColor,anchor=base,inner sep=0pt, outer sep=0pt, scale=  1.00] at (231.65, 39.60) {${10}^{-3}$};

      \node[text=drawColor,anchor=base,inner sep=0pt, outer sep=0pt, scale=  1.00] at (298.24, 39.60) {${10}^{-2}$};

      \node[text=drawColor,anchor=base,inner sep=0pt, outer sep=0pt, scale=  1.00] at (364.83, 39.60) {${10}^{-1}$};

      \node[text=drawColor,anchor=base,inner sep=0pt, outer sep=0pt, scale=  1.00] at (431.41, 39.60) {1};

      \path[draw=drawColor,line width= 0.4pt,line join=round,line cap=round] ( 49.20,105.55) -- ( 49.20,436.60);

      \path[draw=drawColor,line width= 0.4pt,line join=round,line cap=round] ( 49.20,105.55) -- ( 43.20,105.55);

      \path[draw=drawColor,line width= 0.4pt,line join=round,line cap=round] ( 49.20,160.72) -- ( 43.20,160.72);

      \path[draw=drawColor,line width= 0.4pt,line join=round,line cap=round] ( 49.20,215.90) -- ( 43.20,215.90);

      \path[draw=drawColor,line width= 0.4pt,line join=round,line cap=round] ( 49.20,271.07) -- ( 43.20,271.07);

      \path[draw=drawColor,line width= 0.4pt,line join=round,line cap=round] ( 49.20,326.25) -- ( 43.20,326.25);

      \path[draw=drawColor,line width= 0.4pt,line join=round,line cap=round] ( 49.20,381.42) -- ( 43.20,381.42);

      \path[draw=drawColor,line width= 0.4pt,line join=round,line cap=round] ( 49.20,436.60) -- ( 43.20,436.60);

      \node[text=drawColor,rotate= 90.00,anchor=base,inner sep=0pt, outer sep=0pt, scale=  1.00] at ( 34.80,105.55) {30};

      \node[text=drawColor,rotate= 90.00,anchor=base,inner sep=0pt, outer sep=0pt, scale=  1.00] at ( 34.80,160.72) {40};

      \node[text=drawColor,rotate= 90.00,anchor=base,inner sep=0pt, outer sep=0pt, scale=  1.00] at ( 34.80,215.90) {50};

      \node[text=drawColor,rotate= 90.00,anchor=base,inner sep=0pt, outer sep=0pt, scale=  1.00] at ( 34.80,271.07) {60};

      \node[text=drawColor,rotate= 90.00,anchor=base,inner sep=0pt, outer sep=0pt, scale=  1.00] at ( 34.80,326.25) {70};

      \node[text=drawColor,rotate= 90.00,anchor=base,inner sep=0pt, outer sep=0pt, scale=  1.00] at ( 34.80,381.42) {80};

      \node[text=drawColor,rotate= 90.00,anchor=base,inner sep=0pt, outer sep=0pt, scale=  1.00] at ( 34.80,436.60) {90};

      \path[draw=drawColor,line width= 0.4pt,line join=round,line cap=round] ( 49.20, 61.20) --
      (480.69, 61.20) --
      (480.69,456.69) --
      ( 49.20,456.69) --
      ( 49.20, 61.20);
    \end{scope}
\begin{scope}
\path[clip] (  0.00,  0.00) rectangle (505.89,505.89);
\definecolor[named]{drawColor}{rgb}{0.00,0.00,0.00}

\node[text=drawColor,anchor=base,inner sep=0pt, outer sep=0pt, scale=  1.00] at (264.94, 15.60) {Average computation time [s]};

\node[text=drawColor,rotate= 90.00,anchor=base,inner sep=0pt, outer sep=0pt, scale=  1.00] at ( 10.80,258.94) {Relative deviation [\%]};
\end{scope}
\begin{scope}
\path[clip] ( 49.20, 61.20) rectangle (480.69,456.69);
\definecolor[named]{drawColor}{rgb}{0.00,0.00,0.00}

\node[text=drawColor,anchor=base west,inner sep=0pt, outer sep=0pt, scale=  1.00] at (388.07,443.87) {$M_1$};

\node[text=drawColor,anchor=base west,inner sep=0pt, outer sep=0pt, scale=  1.00] at (189.33,122.96) {$\overline{L}_1^a$};

\node[text=drawColor,anchor=base west,inner sep=0pt, outer sep=0pt, scale=  1.00] at (183.09,152.24) {$L_1$};

\node[text=drawColor,anchor=base west,inner sep=0pt, outer sep=0pt, scale=  1.00] at (234.37,141.25) {$\overline{L}_2$};

\node[text=drawColor,anchor=base west,inner sep=0pt, outer sep=0pt, scale=  1.00] at ( 92.32,133.44) {$LC_1$};

\node[text=drawColor,anchor=base west,inner sep=0pt, outer sep=0pt, scale=  1.00] at ( 78.70,191.69) {$LC_2$};

\node[text=drawColor,anchor=base west,inner sep=0pt, outer sep=0pt, scale=  1.00] at (109.22,316.61) {$LC_3$};

\node[text=drawColor,anchor=base west,inner sep=0pt, outer sep=0pt, scale=  1.00] at (389.36,133.02) {$M_2$};

\node[text=drawColor,anchor=base west,inner sep=0pt, outer sep=0pt, scale=  1.00] at (416.52, 77.68) {$M_3$};

\node[text=drawColor,anchor=base west,inner sep=0pt, outer sep=0pt, scale=  1.00] at (265.21,150.88) {$R\mid\mid C_\textrm{max}$};
\definecolor[named]{drawColor}{rgb}{0.83,0.83,0.83}

\path[draw=drawColor,line width= 0.4pt,dash pattern=on 1pt off 3pt ,line join=round,line cap=round] ( 98.48, 61.20) -- ( 98.48,456.69);

\path[draw=drawColor,line width= 0.4pt,dash pattern=on 1pt off 3pt ,line join=round,line cap=round] (231.65, 61.20) -- (231.65,456.69);

\path[draw=drawColor,line width= 0.4pt,dash pattern=on 1pt off 3pt ,line join=round,line cap=round] (364.83, 61.20) -- (364.83,456.69);

\path[draw=drawColor,line width= 0.4pt,dash pattern=on 1pt off 3pt ,line join=round,line cap=round] ( 49.20,105.55) -- (480.69,105.55);

\path[draw=drawColor,line width= 0.4pt,dash pattern=on 1pt off 3pt ,line join=round,line cap=round] ( 49.20,160.72) -- (480.69,160.72);

\path[draw=drawColor,line width= 0.4pt,dash pattern=on 1pt off 3pt ,line join=round,line cap=round] ( 49.20,215.90) -- (480.69,215.90);

\path[draw=drawColor,line width= 0.4pt,dash pattern=on 1pt off 3pt ,line join=round,line cap=round] ( 49.20,271.07) -- (480.69,271.07);

\path[draw=drawColor,line width= 0.4pt,dash pattern=on 1pt off 3pt ,line join=round,line cap=round] ( 49.20,326.25) -- (480.69,326.25);

\path[draw=drawColor,line width= 0.4pt,dash pattern=on 1pt off 3pt ,line join=round,line cap=round] ( 49.20,381.42) -- (480.69,381.42);

\path[draw=drawColor,line width= 0.4pt,dash pattern=on 1pt off 3pt ,line join=round,line cap=round] ( 49.20,436.60) -- (480.69,436.60);
\end{scope}
\end{tikzpicture}
  \caption{Comparison of lower bounds.}
  \label{fig:lowerbounds}
\end{figure}

Figure~\ref{fig:lowerbounds} shows the average relative deviation in percent from the best known value and the
average computation time over all $320$ instances. Looking at the models, the lower bound of $M_2$ is
significantly better than $M_1$, and the addition of the continuity constraints improves the relative
deviation by another $10$\%, yielding the best lower bound overall. The computation time of the three models
is comparable, with $M_3$ being slower than the other two models. The linear relaxation of $R\mid\mid
C_\mathrm{max}$ is slightly worse that $M_2$, but two orders of magnitude faster. The bounds $L_1$ and
$\overline{L}^a_1$ achieve about the same quality an order of magnitude faster than $R\mid\mid
C_\mathrm{max}$. The lower bounds from the relaxation to SALBP are weaker than most of the other lower bounds,
except $LC_1$, but another order of magnitude faster.

For the branch-and-bound we chose to use the lower bounds from the relaxation to SALBP and $\overline{L}_1^a$,
since the other bounds are too costly to be applied at every node of the branch-and-bound tree. We include all
of the faster bounds, since they yield complementary results. In particular $LC_1$ obtains the best bound in
average at the root node, but is less effective during the search.

\subsection{Comparison of MIP models}

We next compare the performance of the new MIP models with that of model $M_1$. Table~\ref{tab:model:results}
shows the average number of nodes and the average computation time needed to solve the instances to optimality
for the $16$ groups with a low number of tasks. The instance groups Tonge and Wee-Mag with a high number of
tasks are not shown, since none of them could be solved to optimality within an hour.

\begin{table}[ht!]
  \scriptsize
  \caption{Comparison of MIP models $M_1$, $M_2$, and $M_3$ on instances Roszieg and Heskia.}
  \begin{center}
    \begin{tabular}{lcccrrr@{\hspace{1em}}rrr} \noalign{\hrule height 1pt}
                         &                      &                     &        & \multicolumn{6}{c}{Model}                                                                                                        \\
                         &                      &                     &        & \multicolumn{2}{c}{$M_1$} & \multicolumn{2}{c}{$M_2$}  & \multicolumn{2}{c}{$M_3$}                                               \\ \cline{5-6}\cline{7-8}\cline{9-10}
      Instance           & $|W|$                & Var                 & Inf    & Nodes                     & \multicolumn{1}{c}{$t$(s)} & Nodes & \multicolumn{1}{c}{$t$(s)} & Nodes & \multicolumn{1}{c}{$t$(s)} \\ \noalign{\hrule height 1pt}\noalign{\vspace{7pt}}
\multirow{8}{*}{Roszieg} & \multirow{4}{*}{$4$} & \multirow{2}{*}{L} & $10\%$ & 56.9   & 0.6                        & 2340.4  & 1.1  & 37.8  & 0.7                        \\
                         &                      &                     & $20\%$ & 1.1    & 0.6                        & 936.0   & 0.4  & 11.7  & 0.4                        \\
                         &                      & \multirow{2}{*}{H} & $10\%$ & 156.6  & 0.8                        & 2849.5  & 1.3  & 58.6  & 1.5                        \\
                         &                      &                     & $20\%$ & 82.9   & 0.8                        & 3268.4  & 1.6  & 53.8  & 0.8                        \\\noalign{\vspace{1ex}}
                         & \multirow{4}{*}{$6$} & \multirow{2}{*}{L} & $10\%$ & 2715.0 & 12.1                       & 47176.3 & 52.4 & 249.9 & 4.6                        \\
                         &                      &                     & $20\%$ & 2601.3 & 11.3                       & 36555.7 & 29.0 & 168.7 & 2.3                        \\
                         &                      & \multirow{2}{*}{H} & $10\%$ & 3467.0 & 13.4                       & 83900.5 & 66.2 & 389.0 & 6.3                        \\
                         &                      &                     & $20\%$ & 2785.0 & 11.8                       & 50294.3 & 44.2 & 281.5 & 4.5                        \\\noalign{\vspace{1ex}}\noalign{\hrule height 1pt}\noalign{\vspace{1ex}}
\multirow{8}{*}{Heskia}  & \multirow{4}{*}{$4$} & \multirow{2}{*}{L} & $10\%$ & 0.0    & 0.6                        & 105.2   & 0.2  & 29.8  & 0.3                        \\
                         &                      &                     & $20\%$ & 25.0   & 0.6                        & 198.6   & 0.3  & 37.5  & 0.2                        \\
                         &                      & \multirow{2}{*}{H} & $10\%$ & 65.0   & 0.7                        & 136.2   & 0.2  & 49.0  & 0.3                        \\
                         &                      &                     & $20\%$ & 24.3   & 0.7                        & 130.5   & 0.3  & 45.5  & 0.2                        \\\noalign{\vspace{1ex}}
                         & \multirow{4}{*}{$7$} & \multirow{2}{*}{L} & $10\%$ & 1535.9 & 13.4                       & 1552.2  & 4.6  & 86.8  & 1.0                        \\
                         &                      &                     & $20\%$ & 1174.1 & 11.1                       & 940.8   & 2.2  & 102.4 & 1.0                        \\
                         &                      & \multirow{2}{*}{H} & $10\%$ & 1677.8 & 12.9                       & 735.9   & 2.5  & 115.4 & 1.1                        \\
                         &                      &                     & $20\%$ & 1344.1 & 13.4                       & 663.3   & 2.8  & 151.7 & 1.3                        \\\noalign{\vspace{1ex}}\noalign{\hrule height 1pt}\noalign{\vspace{1ex}}
\multicolumn{4}{l}{Averages}& 1107.0& 6.6& 14486.5&  13.1& 122.1 & 1.7 \\\noalign{\vspace{1ex}}\noalign{\hrule height 1pt}
    \end{tabular}
  \end{center}
  \label{tab:model:results}
\end{table}

Overall model $M_2$ needs significantly more nodes than $M_1$, and is a factor of about two slower. It
executes more nodes per second, and has a better lower bound, but CPLEX is able to apply more cuts for model
$M_1$ at the root, such that in average model $M_2$ has no advantage on the tested instances. However, when
the continuity constraints are applied, model $M_3$ needs significantly less nodes and time compared to model
$M_1$ (confirmed by a Wilcoxon signed rank test with $p<0.01$). The results show that the continuity
constraints are very effective, in particular for a high order strength and for high numbers of workers.

\subsection{Results for the IPBS heuristic}

We compare IPBS with three state of the art heuristic methods for the ALWABP-2, namely the hybrid genetic
algorithm (HGA) of \citet{Moreira2012Simple}, the iterated beam search (IBS) of \citet{Blum.Miralles/2011},
and the iterative genetic algorithm (IGA) of \citet{Mutlu2013}. In preliminary experiments we determined
reasonable parameters for the probabilistic beam search as shown in Table~\ref{tab:parameters:heuristic}. For
the HGA and the IBS we compare in Table~\ref{tab:values:heuristic} the relative deviation from the current
best known value (Gap) and the computation time ($t$), in average for each group of instances and over $20$
replications per instance. We further report the average computation time to find the best value ($t_b$), and
the average relative deviation of the best solution of the $20$ replications ($Gap_b$). The total computation
time of \citet{Blum.Miralles/2011} is always $120$s more than the time to find the best value, and has been
omitted from the table.

\begin{table}
\footnotesize
\caption{Parameters of the probabilistic beam search used in the computational experiments.}
\center
\begin{tabular}{lr}\hline
  Beam width $w$ & 125 \\
  Branching factor $f$ & 5 \\
  Cycle time reduction for interval search $p$ & 0.95 \\
  Minimum search time $t_{min}$ (s) & 6 \\
  Maximum search time $t_{max}$ (s) & 900 \\
  Maximum number of interval searches $r$ & 20 \\\hline
\end{tabular}
\label{tab:parameters:heuristic}
\end{table}

\begin{table}[ht!]
\scriptsize
\caption{Comparison of the proposed heuristic with a hybrid genetic algorithm~\citep{Moreira2012Simple} and an iterated beam search~\citep{Blum.Miralles/2011}.}
  \begin{center}
    \setlength{\tabcolsep}{0.4em}

\begin{tabular}{lllcrrrrrrrrrrr}
\hline
\multicolumn{4}{c}{} & \multicolumn{ 4}{c}{HGA} & \multicolumn{3}{c}{IBS} & \multicolumn{4}{c}{IBPS} \\\cline{5-8}\cline{9-11}\cline{12-15}
Instance & $|W|$ & Var & Inf & \multicolumn{1}{c}{$t$(s)} & \multicolumn{1}{c}{$t_b$(s)} & \multicolumn{1}{c}{$Gap$} & \multicolumn{1}{c}{$Gap_b$} & \multicolumn{1}{c}{$t_b$(s)} & \multicolumn{1}{c}{$Gap$} & \multicolumn{1}{c}{$Gap_b$} & \multicolumn{1}{c}{$t$(s)} & \multicolumn{1}{c}{$t_b$(s)} & \multicolumn{1}{c}{$Gap$} & \multicolumn{1}{c}{$Gap_b$}  \\
\hline
\multirow{ 8}{*}{Roszieg} & \multirow{4}{*}{4}  & \multirow{2}{*}{L} & 10\% & 3.3   & 0.0   & 0.0  & 0.0 & 0.0   & 0.0  & 0.0 & 6.0   & 0.0  & 0.0 & 0.0 \\ 
                          &                     &                     & 20\% & 4.5   & 0.0   & 0.1  & 0.0 & 0.1   & 0.0  & 0.0 & 5.4   & 0.0  & 0.0 & 0.0 \\ 
                          &                     & \multirow{2}{*}{H} & 10\% & 4.0   & 0.0   & 0.0  & 0.0 & 0.1   & 0.0  & 0.0 & 6.0   & 0.0  & 0.0 & 0.0 \\ 
                          &                     &                     & 20\% & 3.4   & 0.0   & 0.0  & 0.0 & 0.0   & 0.0  & 0.0 & 6.0   & 0.0  & 0.0 & 0.0 \\ 
                          & \multirow{4}{*}{6}  & \multirow{2}{*}{L} & 10\% & 3.6   & 0.0   & 0.0  & 0.0 & 0.0   & 0.0  & 0.0 & 6.0   & 0.0  & 0.0 & 0.0 \\ 
                          &                     &                     & 20\% & 4.0   & 0.1   & 1.1  & 1.0 & 0.0   & 0.0  & 0.0 & 6.0   & 0.1  & 0.0 & 0.0 \\ 
                          &                     & \multirow{2}{*}{H} & 10\% & 4.5   & 0.0   & 0.0  & 0.0 & 0.0   & 0.0  & 0.0 & 6.0   & 0.0  & 0.0 & 0.0 \\ 
                          &                     &                     & 20\% & 4.5   & 0.1   & 0.0  & 0.0 & 0.0   & 0.0  & 0.0 & 6.0   & 0.1  & 0.0 & 0.0 \\ \hline
\multirow{ 8}{*}{Heskia}  & \multirow{4}{*}{4}  & \multirow{2}{*}{L} & 10\% & 6.9   & 0.2   & 0.0  & 0.0 & 8.2   & 0.0  & 0.0 & 6.0   & 0.1  & 0.0 & 0.0 \\ 
                          &                     &                     & 20\% & 9.3   & 0.3   & 0.1  & 0.1 & 3.0   & 0.0  & 0.0 & 6.0   & 0.1  & 0.0 & 0.0 \\ 
                          &                     & \multirow{2}{*}{H} & 10\% & 9.2   & 0.3   & 0.0  & 0.0 & 5.6   & 0.0  & 0.0 & 6.0   & 0.1  & 0.0 & 0.0 \\ 
                          &                     &                     & 20\% & 9.5   & 0.5   & 0.3  & 0.0 & 5.2   & 0.0  & 0.0 & 6.0   & 0.2  & 0.0 & 0.0 \\ 
                          & \multirow{4}{*}{7}  & \multirow{2}{*}{L} & 10\% & 8.0   & 0.2   & 0.5  & 0.0 & 1.1   & 0.0  & 0.0 & 5.4   & 0.2  & 0.0 & 0.0 \\ 
                          &                     &                     & 20\% & 7.4   & 0.3   & 0.6  & 0.0 & 2.5   & 0.0  & 0.0 & 4.3   & 0.2  & 0.5 & 0.3 \\ 
                          &                     & \multirow{2}{*}{H} & 10\% & 6.6   & 0.2   & 0.3  & 0.0 & 1.7   & 0.0  & 0.0 & 2.5   & 0.2  & 0.0 & 0.0 \\ 
                          &                     &                     & 20\% & 9.2   & 1.5   & 0.7  & 0.0 & 2.5   & 0.0  & 0.0 & 3.7   & 0.2  & 0.0 & 0.0 \\ \hline

\multirow{ 8}{*}{Tonge}   & \multirow{4}{*}{10} & \multirow{2}{*}{L} & 10\% & 205.7 & 34.4  & 5.9  & 2.4 & 86.4  & 6.7  & 4.8 & 56.2  & 19.7 & 1.9 & 0.9 \\ 
                          &                     &                     & 20\% & 241.2 & 34.9  & 4.2  & 2.4 & 92.2  & 4.6  & 3.3 & 58.9  & 14.8 & 2.3 & 1.0 \\ 
                          &                     & \multirow{2}{*}{H} & 10\% & 391.0 & 98.6  & 4.4  & 1.8 & 160.1 & 5.5  & 3.5 & 89.2  & 26.8 & 1.4 & 0.9 \\ 
                          &                     &                     & 20\% & 347.5 & 56.9  & 4.3  & 2.7 & 171.4 & 4.7  & 3.8 & 91.0  & 21.4 & 2.2 & 1.2 \\ 
                          & \multirow{4}{*}{17} & \multirow{2}{*}{L} & 10\% & 296.9 & 74.0  & 11.3 & 7.8 & 88.0  & 8.2  & 4.7 & 61.2  & 26.2 & 2.8 & 1.9 \\ 
                          &                     &                     & 20\% & 300.0 & 67.1  & 14.3 & 9.2 & 70.5  & 11.5 & 8.6 & 60.5  & 19.4 & 6.9 & 5.6 \\ 
                          &                     & \multirow{2}{*}{H} & 10\% & 446.7 & 129.1 & 10.9 & 5.9 & 124.3 & 8.1  & 5.6 & 96.7  & 31.8 & 3.2 & 2.5 \\ 
                          &                     &                     & 20\% & 469.5 & 105.3 & 10.4 & 7.6 & 156.4 & 8.9  & 5.8 & 107.7 & 31.5 & 4.5 & 3.2 \\ \hline
\multirow{ 8}{*}{Wee-Mag} & \multirow{4}{*}{11} & \multirow{2}{*}{L} & 10\% & 136.9 & 56.9  & 6.6  & 2.3 & 104.9 & 13.9 & 9.9 & 25.1  & 9.4  & 5.8 & 3.8 \\ 
                          &                     &                     & 20\% & 158.8 & 60.1  & 7.6  & 3.3 & 84.9  & 12.0 & 7.8 & 25.1  & 7.7  & 4.6 & 2.9 \\ 
                          &                     & \multirow{2}{*}{H} & 10\% & 248.5 & 115.8 & 8.9  & 3.9 & 160.3 & 12.6 & 9.3 & 37.1  & 12.4 & 5.6 & 3.7 \\ 
                          &                     &                     & 20\% & 245.9 & 112.7 & 7.8  & 2.9 & 143.3 & 13.8 & 9.6 & 36.1  & 14.1 & 4.5 & 2.6 \\ 
                          & \multirow{4}{*}{19} & \multirow{2}{*}{L} & 10\% & 213.9 & 61.4  & 12.5 & 6.2 & 57.1  & 8.1  & 4.1 & 39.9  & 11.6 & 1.4 & 0.0 \\ 
                          &                     &                     & 20\% & 225.6 & 66.1  & 14.0 & 9.2 & 60.3  & 9.8  & 5.4 & 39.0  & 10.4 & 2.6 & 0.8 \\ 
                          &                     & \multirow{2}{*}{H} & 10\% & 283.7 & 97.9  & 15.3 & 9.1 & 71.4  & 11.1 & 6.5 & 38.3  & 11.0 & 4.6 & 3.7 \\ 
                          &                     &                     & 20\% & 288.1 & 108.9 & 13.0 & 7.7 & 90.0  & 10.3 & 5.8 & 41.6  & 11.8 & 3.8 & 3.5 \\ \hline

         \multicolumn{4}{l}{Total averages} &    143.7&     40.1&      4.9&      2.7     &54.7&      4.7&      3.1&          31.0&      8.8&      1.8&      1.2\\\hline 

\end{tabular}
\end{center}
  \label{tab:values:heuristic}
\end{table}

We can see that the problem can be considered well solved for a low number of tasks, since all three methods
find the optimal solution with a few exceptions in less than ten seconds. In six instance groups the IPBS
terminates in less than the minimum search time, since the solution was provably optimal. For instances with a
high number of tasks, IBS produces better solutions for more workers, while the HGA is better on less
workers. IPBS always achieves better results than both methods (confirmed by a Wilcoxon signed rank test with
$p<0.01$). This holds for the averages as well as the best found solutions (except the first instance group of
Wee-Mag, where the best solution of IPBS is slightly worse than that of the HGA). IPBS is also very robust in
the sense that the difference between average and best relative deviations is the smallest of the three
methods. In average over all instances, its solutions are $1.8$\% over the best known values. Since the best
known values are known to be optimal for $307$ of the $320$ instances, the gap is close to optimal.

To compare execution times, we have to consider that the results have been obtained on different machines (for
IBS a PC with a $2.2$ GHz AMD64X2 $4400$ processor and 4GB of main memory, for HGA a PC with a Core 2 Duo
$2.2$ GHz processor in $3$ GB of main memory). A conservative assumption is that their performance is within a
factor of two of each other. Taking this into account, over all instances HGA and IBS have comparable
computation times, and the IPBS is about a factor two faster. This holds for finding the best solution and
also for the total computation time. (Remember that the total computation time of IBS is $120$\,s longer than
the time to find the best solution.) The faster average computation times are mainly due to the instances with
a high number of tasks, for which IPBS scales better. The best solutions are almost always found in less than
$30$ seconds.

For all heuristics, the computation time is significantly less for a low number of tasks, a low number of
workers, and a low order strength. Similarly, the relative deviations are smaller for a low number of tasks
and low order strength. However, the relative deviation does not depend significantly on the number of
workers, except for the HGA, which produces better solution for a low number of workers. (These findings are
confirmed by a Wilcoxon signed rank test at significance level $p<0.01$.) For IBS and IBPS there is an
interaction between the number of workers and the order strength: both produce better solutions for a low
number of workers and a high order strength or \emph{vice versa}.

\begin{table}
\scriptsize
\caption{Comparison of the proposed heuristic with an iterated genetic
  algorithm~\citep{Mutlu2013}.}
  \begin{center}
    \setlength{\tabcolsep}{0.4em}

\begin{tabular}{lllcrrrrrr}
\hline
\multicolumn{4}{c}{} & \multicolumn{3}{c}{IGA} & \multicolumn{3}{c}{IBPS} \\\cline{5-7}\cline{8-10}
Instance & $|W|$ & Var & Inf & \multicolumn{1}{c}{$t_b$(s)} & \multicolumn{1}{c}{$C$} & \multicolumn{1}{c}{$C_b$} & \multicolumn{1}{c}{$t_b$(s)} & \multicolumn{1}{c}{$C$} & \multicolumn{1}{c}{$C_b$}  \\\hline

\multirow{ 8}{*}{Roszieg}    & \multirow{4}{*}{4}  & \multirow{2}{*}{L} & 10\% & 0.3  & 20.1  & 20.1  & 0.0  & 20.1  & 20.1  \\
                             &                     &                    & 20\% & 0.3  & 31.5  & 31.5  & 0.0  & 31.5  & 31.5  \\
                             &                     & \multirow{2}{*}{H} & 10\% & 0.3  & 28.1  & 28.1  & 0.0  & 28.1  & 28.1  \\
                             &                     &                    & 20\% & 0.2  & 28.0  & 28.0  & 0.0  & 28.0  & 28.0  \\
                             & \multirow{4}{*}{6}  & \multirow{2}{*}{L} & 10\% & 0.5  & 9.7   & 9.7   & 0.0  & 9.7   & 9.7   \\
                             &                     &                    & 20\% & 0.5  & 11.0  & 11.0  & 0.1  & 11.0  & 11.0  \\
                             &                     & \multirow{2}{*}{H} & 10\% & 0.5  & 16.0  & 16.0  & 0.0  & 16.0  & 16.0  \\
                             &                     &                    & 20\% & 0.5  & 15.1  & 15.1  & 0.0  & 15.1  & 15.1  \\\hline
\multirow{ 8}{*}{Heskia}     & \multirow{4}{*}{4}  & \multirow{2}{*}{L} & 10\% & 0.3  & 102.3 & 102.3 & 0.1  & 102.3 & 102.3 \\
                             &                     &                    & 20\% & 0.3  & 122.6 & 122.6 & 0.1  & 122.6 & 122.6 \\
                             &                     & \multirow{2}{*}{H} & 10\% & 0.3  & 172.5 & 172.5 & 0.1  & 172.5 & 172.5 \\
                             &                     &                    & 20\% & 0.2  & 171.3 & 171.2 & 0.2  & 171.3 & 171.2 \\
                             & \multirow{4}{*}{7}  & \multirow{2}{*}{L} & 10\% & 0.5  & 34.9  & 34.9  & 0.2  & 34.9  & 34.9  \\
                             &                     &                    & 20\% & 0.5  & 42.6  & 42.6  & 0.2  & 42.8  & 42.7  \\
                             &                     & \multirow{2}{*}{H} & 10\% & 0.5  & 75.2  & 75.2  & 0.1  & 75.2  & 75.2  \\
                             &                     &                    & 20\% & 0.5  & 67.2  & 67.2  & 0.2  & 67.2  & 67.2  \\\hline
\multirow{ 8}{*}{Tonge}      & \multirow{4}{*}{10} & \multirow{2}{*}{L} & 10\% & 47.4 & 94.1  & 93.0  & 19.7 & 92.2  & 91.3  \\
                             &                     &                    & 20\% & 40.5 & 110.2 & 109.3 & 14.8 & 109.1 & 107.8 \\
                             &                     & \multirow{2}{*}{H} & 10\% & 70.8 & 165.2 & 162.4 & 26.8 & 161.7 & 160.8 \\
                             &                     &                    & 20\% & 59.4 & 170.1 & 168.4 & 21.4 & 167.5 & 165.9 \\
                             & \multirow{4}{*}{17} & \multirow{2}{*}{L} & 10\% & 78.0 & 33.1  & 33.1  & 26.2 & 32.5  & 32.2  \\
                             &                     &                    & 20\% & 68.4 & 40.4  & 40.1  & 19.4 & 39.4  & 38.9  \\
                             &                     & \multirow{2}{*}{H} & 10\% & 68.1 & 66.4  & 66.4  & 31.8 & 64.9  & 64.5  \\
                             &                     &                    & 20\% & 78.0 & 64.8  & 64.6  & 31.5 & 63.9  & 63.1  \\\hline
\multirow{ 8}{*}{Wee-Mag}    & \multirow{4}{*}{11} & \multirow{2}{*}{L} & 10\% & 65.7 & 27.4  & 26.7  & 9.4  & 27.6  & 27.1  \\
                             &                     &                    & 20\% & 61.8 & 32.7  & 32.3  & 7.7  & 32.6  & 32.1  \\
                             &                     & \multirow{2}{*}{H} & 10\% & 92.7 & 48.2  & 47.6  & 12.4 & 48.4  & 47.5  \\
                             &                     &                    & 20\% & 81.9 & 46.0  & 45.8  & 14.2 & 46.2  & 45.4  \\
                             & \multirow{4}{*}{19} & \multirow{2}{*}{L} & 10\% & 67.2 & 10.4  & 10.3  & 11.6 & 10.0  & 9.9   \\
                             &                     &                    & 20\% & 67.2 & 12.1  & 12.1  & 10.4 & 11.6  & 11.4  \\
                             &                     & \multirow{2}{*}{H} & 10\% & 68.1 & 18.5  & 18.2  & 11.0 & 17.9  & 17.7  \\
                             &                     &                    & 20\% & 77.4 & 18.4  & 18.0  & 11.8 & 17.7  & 17.7  \\\hline
\multicolumn{4}{l}{Averages}                                                    & 34.3 & 59.6  & 59.3  & 8.8  & 59.1  & 58.8  \\\hline 
\end{tabular}
\end{center}
  \label{tab:values:heuristic:mutlu}
\end{table}

Since for the IGA no detailed results are available, we compare in Table~\ref{tab:values:heuristic:mutlu} with
the summarized values reported by \citet{Mutlu2013}: the average cycle time ($C$), the average cycle time of
the best found solution ($C_b$), and the average computation time to find the best value ($t_b$). The values
are again averages for all groups of instances, but over only $10$ replications for the IGA. The results for
our method are the same as in Table~\ref{tab:values:heuristic} but in absolute values. Note that this
evaluation may mask large deviations in instances with low cycle times and overestimate small deviations for
high cycle times.

As the other methods, the IGA solves the small instances optimally, but not the larger ones. Compared to our
method, its average performance is worse except for three groups of wee-mag with a low number of workers,
where the average cycle time is about $0.2$ lower. The comparison is similar for the best found values, where
the IGA is better by $0.4$ in a single group. In average over all large instances our method produces a cycle
time of about $1$ unit less.

The execution times of the two methods are comparable. The results of \citet{Mutlu2013} have been obtained on
a Intel Core 2 Duo T5750 processor running at $2.0$~GHz, whose performance is within a factor of three from
our machine. Taking this into account, our methods find the best value about $50$\% faster.

In summary, the results show that IPBS can compete with and often outperforms the other methods in solution
quality as well as computation time. The difference to the other methods is smallest for the large instances
with a low order strength and a low number of workers. IPBS in general is very robust over the entire set of
instances.

\subsection{Results for the branch-and-bound algorithm}

We evaluated the branch-and-bound algorithm on the same $320$ test instances. For the tests, IPBS was used to
produce an initial heuristic solution. It was made deterministic by fixing a random seed of $42$ and
configured with a minimum search time of $0\,$s and a maximum search time of $|T||W|/10\,$s. During the search
the number of iterations of the ascent direction method to compute $L_1$ has been limited to $50$, and the
number of iterations for the subgradient optimization to compute $L_2$ to $20$.

The only other branch-and-bound algorithm in the literature proposed by \citet{miralles2008branch} for the
ALWABP-2 has been found inferior to model $M_1$ by \citet{Chaves2009c} in tests with CPLEX (version 10.1). We
therefore limit our comparison to the MIP models. We first compare our approach to CPLEX on the best model
$M_3$ on the instances with a low number of tasks in Table~\ref{tab:results:branchandbound:small}. In
Table~\ref{tab:results:bigger} we then present the results of the branch-and-bound algorithm with a time limit
of one hour on the larger instances. CPLEX is not able to solve any of the models on the instances with a high
number of tasks within this time limit.

\begin{table}
  \scriptsize
  \caption{Comparison of model $M_3$ to the branch-and-bound algorithm on instances with a low number of workers.}
  \begin{center}
    \begin{tabular}{l c c c r r c r r} \noalign{\hrule height 1pt}
                         &                      &                     &        & \multicolumn{2}{c}{\textbf{Model $M_3$}} && \multicolumn{2}{c}{\textbf{B\&B}} \\ \cline{5-6}\cline{8-9}
      Instance           & $|W|$                & Var                 & Inf    & $t$ (s)   & Nodes && $t$ (s) & Nodes            \\ \noalign{\hrule height
1pt}\noalign{\vspace{1ex}}
\multirow{8}{*}{Roszieg} & \multirow{4}{*}{$4$} & \multirow{2}{*}{L} & $10\%$ & 0.7                                      & 37.8  && 0.2    & 34.0             \\
                         &                      &                     & $20\%$ & 0.4                                      & 11.7  && 0.2    & 16.1             \\
                         &                      & \multirow{2}{*}{H} & $10\%$ & 1.5                                      & 58.6  && 0.3    & 44.8             \\
                         &                      &                     & $20\%$ & 0.8                                      & 53.8  && 0.2    & 37.9             \\\noalign{\vspace{1ex}}
                         & \multirow{4}{*}{$6$} & \multirow{2}{*}{L} & $10\%$ & 4.6                                      & 249.9 && 0.7    & 126.5            \\
                         &                      &                     & $20\%$ & 2.3                                      & 168.7 && 0.7    & 77.9             \\
                         &                      & \multirow{2}{*}{H} & $10\%$ & 6.3                                      & 389.0 && 0.8    & 208.1            \\
                         &                      &                     & $20\%$ & 4.5                                      & 281.5 && 0.8    & 130.5            \\\noalign{\vspace{1ex}}\noalign{\hrule height 1pt}\noalign{\vspace{1ex}}
\multirow{8}{*}{Heskia}  & \multirow{4}{*}{$4$} & \multirow{2}{*}{L} & $10\%$ & 0.3                                      & 29.8  && 0.6    & 35.3             \\
                         &                      &                     & $20\%$ & 0.2                                      & 37.5  && 0.6    & 40.8             \\
                         &                      & \multirow{2}{*}{H} & $10\%$ & 0.3                                      & 49.0  && 0.7    & 54.6             \\
                         &                      &                     & $20\%$ & 0.2                                      & 45.5  && 0.7    & 64.8             \\\noalign{\vspace{1ex}}
                         & \multirow{4}{*}{$7$} & \multirow{2}{*}{L} & $10\%$ & 1.0                                      & 86.8  && 2.5    & 20.2             \\
                         &                      &                     & $20\%$ & 1.0                                      & 102.4 && 2.9    & 23.2             \\
                         &                      & \multirow{2}{*}{H} & $10\%$ & 1.1                                      & 115.4 && 2.4    & 13.5             \\
                         &                      &                     & $20\%$ & 1.4                                      & 151.7 && 3.1    & 17.3             \\\noalign{\vspace{1ex}}\noalign{\hrule height 1pt}\noalign{\vspace{1ex}}
\multicolumn{4}{l}{Averages}& 1.7& 116.8& &  1.1&  59.1 \\\noalign{\vspace{1ex}}\noalign{\hrule height 1pt}
    \end{tabular}
  \end{center}
  \label{tab:results:branchandbound:small}
\end{table}

Table~\ref{tab:results:branchandbound:small} shows the average solving time and the average number of nodes in
the branch-and-bound tree for all instance groups with a low number of workers. On these instances both
methods have a similar performance, solving all instances in a few seconds, and are even competitive with the
heuristic methods. In most cases the branch-and-bound algorithm needs fewer nodes than CPLEX, except for five
groups with a low number of workers. Computation times are also comparable, although the time of the
branch-and-bound algorithm is dominated by the initial heuristic.

\begin{table}
  \scriptsize
  \caption{Results of the branch-and-bound algorithm on instances with a high number of workers.}
  \begin{center}
    \setlength{\tabcolsep}{0.4em}
    \begin{tabular}{l c c c r r r r r r } \noalign{\hrule height 1pt}
\multicolumn{1}{c}{Instance} & $|W|$                 & Var                 & Inf    & Opt. & Prov. & \multicolumn{1}{c}{$t$(s)} & \multicolumn{1}{c}{$Gap$} & \multicolumn{1}{c}{$C$} \\ \noalign{\hrule height 1pt}\noalign{\vspace{1ex}}
\multirow{8}{*}{Tonge}       & \multirow{4}{*}{$10$} & \multirow{2}{*}{L} & $10\%$  & 10 & 10 & 175.7  & 0.0 & 90.6                 \\
                             &                       &                     & $20\%$  & 10 & 10 & 144.1  & 0.0 & 106.7                \\
                             &                       & \multirow{2}{*}{H} & $10\%$  & 10 & 10 & 406.8  & 0.0 & 159.3                \\
                             &                       &                     & $20\%$  & 10 & 10 & 213.4  & 0.0 & 163.9                \\\noalign{\vspace{1ex}}
                             & \multirow{4}{*}{$17$} & \multirow{2}{*}{L} & $10\%$  & 10 & 10 & 775.5  & 0.0 & 31.6                 \\
                             &                       &                     & $20\%$  & 10 & 9  & 928.6  & 0.0 & 36.9                 \\
                             &                       & \multirow{2}{*}{H} & $10\%$  & 7  & 7  & 1453.0 & 0.7 & 63.5                 \\
                             &                       &                     & $20\%$  & 10 & 10 & 1211.7 & 0.0 & 61.2                 \\\noalign{\vspace{1ex}}\noalign{\hrule height 1pt}\noalign{\vspace{1ex}}
\multirow{8}{*}{Wee-mag}     & \multirow{4}{*}{$11$} & \multirow{2}{*}{L} & $10\%$  & 5  & 3  & 3102.1 & 2.2 & 26.7                 \\
                             &                       &                     & $20\%$  & 5  & 2  & 3504.9 & 2.3 & 31.9                 \\
                             &                       & \multirow{2}{*}{H} & $10\%$  & 4  & 2  & 3051.6 & 2.4 & 46.9                 \\
                             &                       &                     & $20\%$  & 4  & 4  & 2853.2 & 1.9 & 45.2                 \\\noalign{\vspace{1ex}}
                             & \multirow{4}{*}{$19$} & \multirow{2}{*}{L} & $10\%$  & 2  & 0  & 3600.0 & 2.2 & 10.1                 \\
                             &                       &                     & $20\%$  & 5  & 2  & 3151.5 & 0.9 & 11.4                 \\
                             &                       & \multirow{2}{*}{H} & $10\%$  & 4  & 3  & 2732.1 & 3.7 & 17.7                 \\
                             &                       &                     & $20\%$  & 4  & 3  & 3032.5 & 2.9 & 17.6                 \\\noalign{\vspace{1ex}}\noalign{\hrule height 1pt}\noalign{\vspace{1ex}}
\multicolumn{4}{l}{Totals/Averages}& 111& 95& 1896.0 & 1.20  \\\noalign{\vspace{1ex}}\noalign{\hrule height 1pt}
    \end{tabular}
  \end{center}
  \label{tab:results:bigger}
\end{table}

Table~\ref{tab:results:bigger} shows the results of the branch-and-bound algorithm on the larger instances. We
report the number of optimal solutions found (Opt) and the number of solutions proven to be optimal (Prov),
the average computation time ($t$), the average relative deviation from the best known value ($Gap$), and the
average cycle time for each group of instances ($C$). 

In about $70$\% of the instances the optimal solution was found, and about $60$\% of the solutions could be
proven to be optimal within the time limit. All except four instances with a high order strength were
solved. The average relative deviation over all $320$ instances is $0.60$\%, about one third of the average
case of the best heuristic.

As expected, the solution times are higher than those of the heuristic methods but for the instances with a
high order strength only about an order of magnitude, in average. The solving time depends mainly on the
number of tasks, the number of workers, and the order strength (as confirmed by a Kruskal-Wallis test followed
by Wilcoxon signed rank post hoc tests at significance level $p<0.01$). The instances with a high order
strength or a low number of workers are easier to solve, because the reduction rules are more effective.

\section{Conclusion}
\label{sec:conclusion}

We have presented a new MIP model, a heuristic search procedure and an exact algorithm for solving the
Assembly Line Worker Assignment and Balancing Problem of type 2.  The new MIP model shows the importance of
including continuity constraints in this type of problem, and its linear relaxation gives the current best
lower bound for the problem. The proposed heuristic IPBS is competetive with the current best methods, often
outperforms them in computation time and solution quality, and shows a robust performance over the complete
set of $320$ test instances. Finally, the branch-and-bound method can solve instances with a low number of
tasks in a few seconds, and was able to optimally solve $95$ of the $160$ instances with a high number of
tasks for the first time.

With respect to the problem, constraints that enforce continuity have shown to be the most effective way of
strengthening the lower bounds in the models as well as the heuristic and exact algorithm. Besides the size of
the instance, the number of workers and the order strength has the strongest influence on the problem
difficulty. All methods are able to solve instances with a high order strength better. This also holds for the
branch-and-bound algorithm on instances with a low number of workers.

Our results show that assembly lines with heterogeneous workers can be balanced robustly and close to optimal
for problems of sizes of about $75$ tasks and $20$ workers. Problems of this size arise, for example, in
Sheltered Work Centers for Disabled, and we hope that these methods will contribute to a better integration of
persons with disabilities in the labour market. A very interesting future line of research in this context may
be the integration of persons with disabilities into larger assembly lines with regular workers.

\bibliographystyle{plainnat}
\bibliography{marcus}

\end{document}